%% file: ijcai25.tex
\documentclass{article}
\usepackage{ijcai25}
\usepackage{amsthm,amsmath,amssymb}
\usepackage{mathrsfs}
\usepackage{times}
\usepackage{soul}
\usepackage{url}
\usepackage[hidelinks]{hyperref}
\usepackage[utf8]{inputenc}
\usepackage[small]{caption}
\usepackage{graphicx}
\usepackage{amsmath}
\usepackage{amsthm}
\usepackage{booktabs}
\usepackage{algorithm}
\usepackage{algorithmic}
\usepackage{multirow}
\usepackage[switch]{lineno}
\usepackage{xcolor} 
\usepackage{amssymb}
\usepackage{subcaption}
\usepackage{enumitem}
\usepackage{booktabs}

\newcommand\blfootnote[1]{%
\begingroup
\renewcommand\thefootnote{}\footnote{#1}%
\addtocounter{footnote}{-1}%
\endgroup
}

\title{BalanceBenchmark: A Survey for Multimodal Imbalance Learning}
\author{
Shaoxuan Xu$^{1,\dagger}$
\and
Menglu Cui$^{2,\dagger}$\and
Chengxiang Huang$^{3}$ \and
Hongfa Wang$^{4,5}$ \And
Di Hu$^{1,*}$
\affiliations
$^1$Gaoling School of Artificial Intelligence, Renmin University of China, Beijing, China\\
$^2$Shanghai University of Finance and Economics, Shanghai, China\\
$^3$Beijing University of Posts and Telecommunications, Beijing, China\\
$^4$Tencent Data Platform, Shenzhen, China\\
$^5$Tsinghua Shenzhen International Graduate School, Shenzhen, China\\
\emails
\{xushaoxuan20040225, dihu\}@ruc.edu.cn,
Louise158@stu.sufe.edu.cn,
huangchengxiang2021@bupt.edu.cn,
hongfawang@tencent.com
}
\begin{document}
\maketitle
\begin{abstract}
Multimodal learning has gained attention for its capacity to integrate information from different modalities. However, it is often hindered by the \textit{multimodal imbalance problem}, where certain modality dominates while others remain underutilized. Although recent studies have proposed various methods to alleviate this problem, they lack comprehensive and fair comparisons. In this paper, we systematically categorize various mainstream multimodal imbalance algorithms into four groups based on the strategies they employ to mitigate imbalance. To facilitate a comprehensive evaluation of these methods, we introduce \textbf{BalanceBenchmark}, a benchmark including multiple widely used multidimensional datasets and evaluation metrics from three perspectives: performance, imbalance degree, and complexity. To ensure fair comparisons, we have developed a modular and extensible toolkit that standardizes the experimental workflow across different methods. Based on the experiments using  BalanceBenchmark, we have identified several key insights into the characteristics and advantages of different method groups in terms of performance, balance degree and computational complexity. We expect such analysis could inspire more efficient approaches to address the imbalance problem in the future, as well as foundation models. The code of the toolkit is available at \url{https://github.com/GeWu-Lab/BalanceBenchmark}.
\blfootnote{\noindent
\textsuperscript{$\dagger$}Equal contribution. 
\textsuperscript{*}Corresponding author.
}
\end{abstract}
\input{Intro/introduction}

\input{Preliminaries/Preliminaries}
\input{Taxonomy/Taxonomy}
\input{Methods/Methods}
\input{Datasets/Datasets}
\input{Experiments/Experiments}

\input{Analysis/Analysis_Conclusions}
\clearpage
\bibliographystyle{named}
\bibliography{ijcai25}

\end{document}

%% file: Intro/introduction.tex
\section{Introduction}
Humans perceive the real world through multiple sensory modalities, such as visual, auditory, and haptic inputs. This rich interplay of modalities has driven extensive research into multimodal learning \cite{Bal_MM}. However, recent studies have identified a critical challenge in this field: the multimodal imbalance problem, where certain modalities disproportionately dominate the behavior of multimodal models \cite{Peng_2022_CVPR}, which impairs the integration and utilization of information across different modalities. To address this issue, researchers have proposed a wide range of approaches aimed at mitigating this problem, which has gained increasing attention \cite{Peng_2022_CVPR,Gblending_Wang,MMCosine_Xu,AGM_Li,CML_Ma,Greedy_Wu,MMPareto_Wei,PMR_Fan,UMT_Du,ReconBoost_Huang}.

\begin{figure*}[t]
    \centering
    \includegraphics[width=0.75\textwidth]{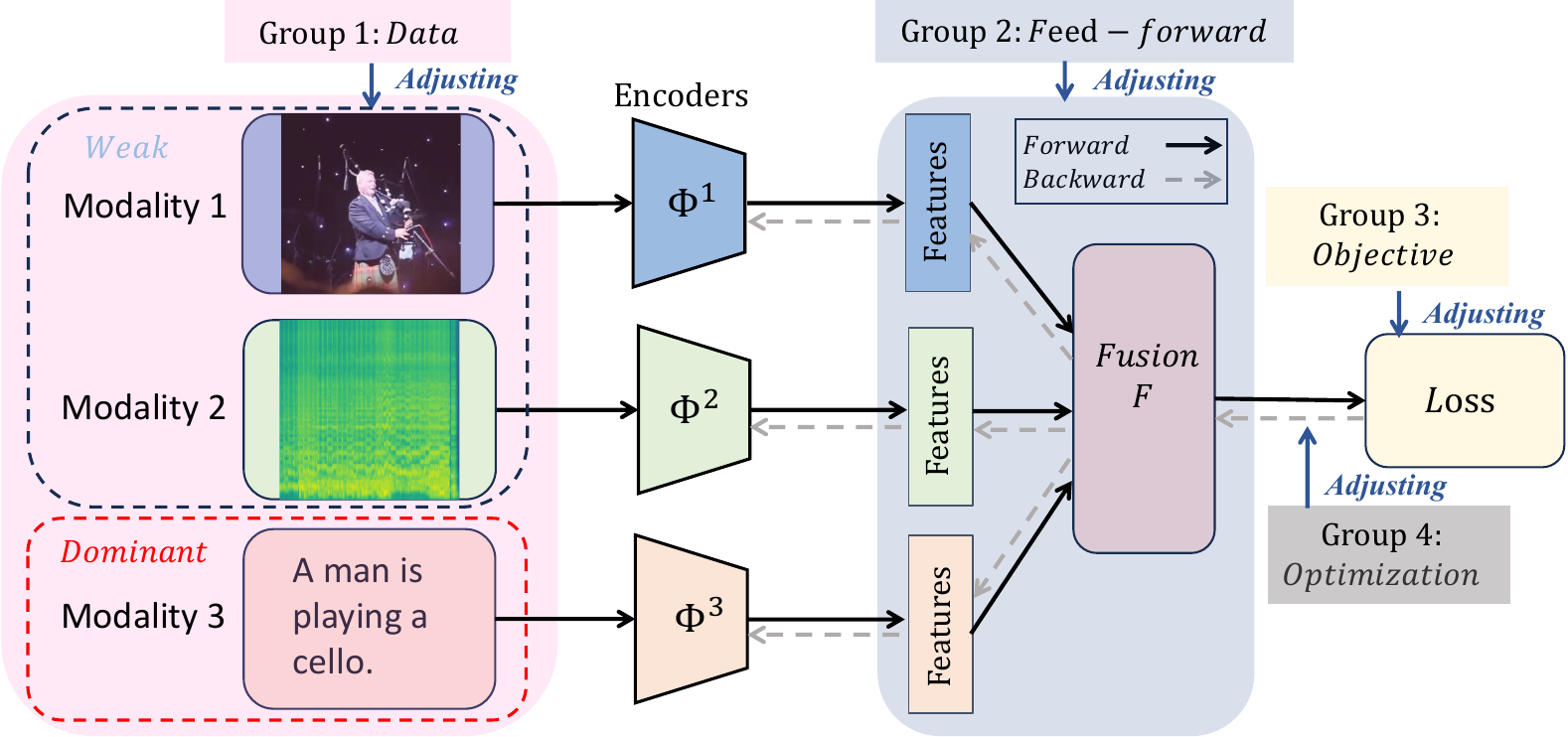}
    \caption{The general framework of multimodal imbalance learning. Group 1 applies adjustments during data processing. Group 2 modifies the fusion module in the feed-forward propagation. Group 3 adapts learning objectives, and Group 4 focuses on optimization adjustments.}
    \label{fig:method_graph}
    \vspace{-15pt}
\end{figure*}

As shown in Figure \ref{fig:method_graph}, these methods employ different strategies within a general framework to address the imbalance problem. However, the lack of comprehensive and fair comparisons makes it difficult to objectively evaluate their effectiveness. This challenge arises from three key issues:
\textit{Firstly, the lack of diverse and representative datasets.} Most multimodal imbalance algorithms have only been evaluated on a limited number of datasets, which do not adequately capture variations in modality counts, modality type, and imbalance degree. This limitation restricts the assessment of a method’s generalizability across real-world scenarios.
\textit{Secondly, the lack of diverse evaluation metrics.} Existing evaluation metrics primarily emphasize performance improvements while overlooking other critical perspectives such as modality imbalance and computational complexity. Moreover, they fail to explore the relationship between model performance and modality imbalance.
\textit{Thirdly, the lack of a standardized experimental workflow.} The absence of a standardized experimental workflow leads to inconsistent experimental setting. Different studies adopt varying experimental settings, making direct comparisons between methods unreliable. 

Given the challenges and limitations discussed above, we first review recent advancements in multimodal imbalance learning and systematically categorize existing methods based on their underlying principles. We then introduce \textbf{BalanceBenchmark}, a comprehensive evaluation framework designed to assess 17 representative methods across seven multidimensional datasets. These datasets cover a wide range of modality combinations, including audio-visual, text-visual, optical flow-RGB, and audio-visual-text modalities, with sample sizes varying from 10K to 200K. Our evaluation metrics include accuracy and F1-score to measure model performance. To quantify modality imbalance, we use Shapley value \cite{shapley:book1952}, which evaluates the contribution of each modality to the final prediction. Additionally, we assess model complexity using floating point operations (FLOPs), which reflect the computational cost required for training. To ensure fair comparisons, we provide \textbf{BalanceMM}, a modular and extensible toolkit designed to standardize the experimental workflow for evaluating different methods. Based on comprehensive experiments, we find that no existing method achieves a satisfactory balance between performance and computational cost. Meanwhile, greater balance between modalities does not guarantee better performance.

Overall, our main contributions are summarized as follows:
\begin{itemize}
\item \textit{Firstly,} we present a systematic taxonomy of existing methods categorized by their strategies for mitigating the imbalance problem, along with a benchmark, \textbf{BalanceBenchmark}, which includes multidimensional datasets and comprehensive evaluation metrics.
\item \textit{Secondly,} we introduce a modular toolkit \textbf{BalanceMM}, which standardizes the experimental workflow for evaluating different methods.
\item \textit{Thirdly,} we use BalanceMM to conduct comprehensive experiments and analyses on existing methods, offering insights into future research directions.
\end{itemize}

%% file: Preliminaries/Preliminaries.tex
\section{Multimodal imbalance learning}
\label{sec:pre}

\input{Preliminaries/Imbalance}

\input{Preliminaries/Problem_define}

%% file: Preliminaries/Imbalance.tex
\label{subsec:imbalance}
Multimodal learning aims to leverage diverse information from different modalities to enhance model performance \cite{Bal_MM}. However, recent studies have revealed the multimodal imbalance problem, where models tend to over-rely on some modalities while underutilizing others \cite{OGM_CVPR}. This imbalance leads to suboptimal exploitation of the available multimodal information.
\begin{figure}[t]
  \centering
  \begin{subfigure}[t]{0.96\linewidth}
    \centering
    \includegraphics[width=0.9\linewidth]{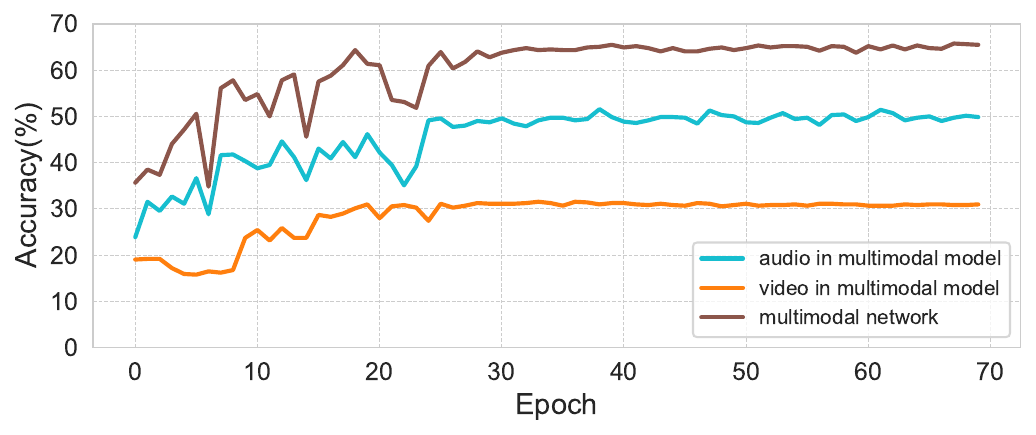} 
    \caption{Multimodal model performance.}
    \label{fig:multimodal_comparison}
  \end{subfigure}
  \vspace{-5pt} 
  \begin{subfigure}[t]{0.48\linewidth} 
    \centering
    \includegraphics[width=0.9\linewidth]{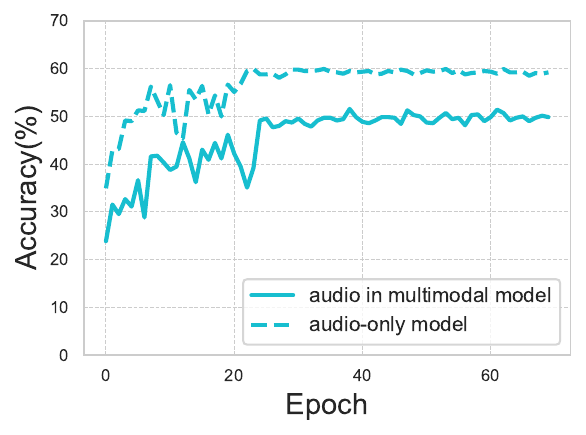}
    \caption{Audio performance gap}
    \label{fig:audio_comparison}
  \end{subfigure}%
  \hfill
  \begin{subfigure}[t]{0.48\linewidth}
    \centering
    \includegraphics[width=0.9\linewidth]{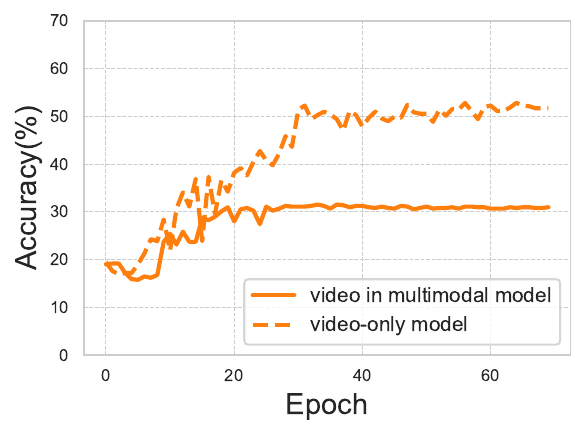}
    \caption{Video performance gap}
    \label{fig:visual_comparison}
  \end{subfigure}
  \caption{\textbf{(a).}  Performance comparison of the multimodal model with its unimodal counterparts on CREMA-D. \textbf{(b).} Performance gap between audio modality within multimodal model and audio-only model on CREMA-D. \textbf{(c).} Performance gap between video modality within multimodal model and video-only model on CREMA-D.}
  \vspace{-10pt} 
  \label{fig:imbalance illustration}
\end{figure}

We consider a general multimodal learning framework for the illustration of imbalance phenomenon. Let $D_{train} = \{(x_k,y_k)\}_{k=1}^{N}$ denote the multimodal training dataset. Each sample $x_k = (x_k^1,x_k^2, \dots,x_k^m)$ consists of $m$ modalities, and $y_k \in \{1,2,\dots,H\}$ denotes the corresponding class label from $H$ classes.
In a multimodal model, each modality uses its own encoder $\Phi^i(\theta^i,\cdot)$ with parameters $\theta_i$. For simplicity, we write it as $\Phi^i$.
As the example, we take the most widely used vanilla fusion method, concatenation. Then the logits output of the multimodal model can be written as :
\begin{equation} \label{equ:1}
    f(x_k) = W[\Phi_k^1;\Phi_k^2;\cdots;\Phi_k^m] + b,
\end{equation}
where $W \in \mathbb{R}^{H \times \sum_i^m d_{\Phi^i}}$ and $b \in \mathbb{R}^{H}$ are the parameters of the last linear classifier. $W$ can be represented as the combination of $m$ blocks: $[W^1,W^2,\cdots,W^m]$. The equation can be rewritten as: 
\begin{equation}
    f(x_k) = \sum_{i=1}^m W^i \cdot \Phi_k^i + b    
\end{equation}
We denote $\hat{y}_k$ as the classification result of $x_k$ by logits output $f(x_k)$. Then the cross-entropy loss is calculated as:
\begin{equation}
    L = \frac{1}{N}\sum_{k=1}^N\ell(\hat{y}_k, y_k),
\end{equation}
where $\ell$ denotes cross-entropy loss. 

With the Gradient Descent optimization method, $W^i$ and the parameters of encoder $\Phi^i$ are updated as:
\begin{equation}
    W^i_{t+1} = W^i_t - \eta\frac{1}{N}\sum_{k=1}^N \frac{\partial \mathcal{L}}{\partial f(x_k)}\Phi_k^i,
\end{equation}
\begin{equation}
    \theta^i_{t+1} = \theta^i_t - \eta\frac{1}{N}\sum_{k=1}^N \frac{\partial \mathcal{L}}{\partial f(x_k)}\frac{\partial(W^i_t \cdot \Phi_k^i)}{\partial \theta^i_t},
\end{equation}
where $\eta$ is the learning rate. 
According to the gradient update equations, the term $\frac{\partial \mathcal{L}}{\partial f(x_k)}$ can be further derived as:
\begin{equation}
    \frac{\partial \mathcal{L}}{\partial f(x_k)_{\hat{y}_k}} = \frac{e^{\sum_{i=1}^m W^i\cdot\Phi_k^i + b_{\hat{y}_k}}}{\sum_{h=1}^H e^{\sum_{i=1}^m W^i\cdot\Phi_k^i + b_h}} - \mathbf{1}_{{\hat{y}_k}=y_k}
\end{equation}
Based on the gradient update equations, recent studies have revealed that when one modality has better performance, its contribution $W^i \cdot \Phi_k^i$ dominates the logits output $f(x_k)$. This reduces the magnitude of $\frac{\partial \mathcal{L}}{\partial f(x_k)}$, as the loss $\mathcal{L}$ already becomes smaller. Consequently, gradients for updating weaker modalities are suppressed, leading to under-optimized representations for them.

To further verify the multimodal imbalance problem, we conduct experiments on CREMA-D dataset \cite{cremad}. It is a widely used audio-video dataset, particularly suitable for studying modality imbalance. We compare three settings: (1) the multimodal model that jointly learns from both audio and video modalities, (2) the unimodal counterparts within this multimodal model, and (3) unimodal models trained using only single modality data. As shown in Figure \ref{fig:imbalance illustration}, while the multimodal model outperforms the unimodal counterparts, both audio and video modalities in the multimodal model performs worse than when trained alone. Besides, the video modality shows a bigger drop in performance, which means weaker modalities are supressed during training. These results align with the previous analysis about the imbalance problem. To alleviate this problem, recent studies have proposed various methods from adjusting the training data distribution to modifying the optimization process.

%% file: Taxonomy/Taxonomy.tex
\section{Taxonomy}
\label{sec:taxonomy}
In this section, we present our taxonomy for mitigating the multimodal imbalance problem based on the strategies for handling modality imbalance. As shown in Table \ref{tab:mm_algorithms}, we categorize these methods into four groups: Data, Feed-forward, Objective and Optimization. We also summarize the different types of imbalance indicator, which different methods use to evaluate the performance of different modalities.

\begin{table*}[t] 
    \centering
    \setlength{\tabcolsep}{5pt}
    \caption{Multimodal imbalance algorithms. \textbf{Adjustment Strategy} refers to different groups of methods in Section \ref{sec:taxonomy}. \textbf{Imbalance Indicator} denotes the metric used to evaluate modality performance. \textbf{Number of Modalities} indicates the maximum number of modalities included in the experiments of the corresponding paper. \textbf{Dataset Domain} refers to the types of modalities included in the corresponding paper.} 
\vspace{-5pt}
\label{tab:mm_algorithms} 
    \begin{tabular}{l|c|c|c|ccc}
        \toprule
        \multicolumn{1}{c}{}            & \multicolumn{1}{c}{Adjustment}       & \multicolumn{1}{c}{Imbalance}         & \multicolumn{1}{c}{Number of}                & \multicolumn{3}{c}{Dataset Domain} \\
        \multicolumn{1}{l}{Method}   & \multicolumn{1}{c}{Strategy}         & \multicolumn{1}{c}{Indicator}    & \multicolumn{1}{c}{Modalities}         & CV & NLP & Audio \\                 
        \midrule
                                                                
        Modality-valuation \cite{Sample_wei}       & Data                                     & Shapley-based Metric                                                     & $2$                                        & \checkmark    &  & \checkmark \\
        MLA \cite{MLA}       & Feed-forward                                     & N/A                                                    & $3$                                          & \checkmark    & \checkmark & \checkmark \\
        OPM \cite{OPM_PAMI}       & Feed-forward                                     & Performance Score                                                    & $3$                                         & \checkmark    & \checkmark & \checkmark \\
        Greedy \cite{Greedy_Wu}       & Feed-forward                                     & Gradient Change                                                     & $2$                                         & \checkmark    &  & \checkmark \\
        AMCo \cite{AMCo}       & Feed-forward                                     & Performance Score                                                     & $3$                                         & \checkmark    & \checkmark & \checkmark \\
        MMCosine \cite{MMCosine_Xu} & Objective                                   & Performance Score                                                       & $2$                                   &     \checkmark  &  & \checkmark \\   
        UMT \cite{UMT_Du}         & Objective                                      & N/A                                                     & $3$                                         & \checkmark    &  & \checkmark \\
        MBSD \cite{MBSD}       & Objective                                     & Performance Score                                                     & $2$                                         & \checkmark    & \checkmark &\\
        CML \cite{CML_Ma}          & Objective                                    & Classification Loss                                                    & $2$                                         & \checkmark    & \checkmark  & \\
        MMPareto \cite{MMPareto_Wei}          & Objective                                    & Performance Score                                                    & $3$                                         & \checkmark    & \checkmark & \checkmark  \\
        GBlending \cite{Gblending_Wang}       & Objective                                     & Classification Loss                                                     & $3$                                          &\checkmark     &   & \checkmark   \\
        LFM \cite{LFM_yang}        & Objective                                      &  N/A                                                              & $3$                                             & \checkmark    & \checkmark  &  \checkmark \\
        OGM \cite{OGM_CVPR}  & Optimazation                                     & Performance Score                                                       & $2$                                          & \checkmark    &  & \checkmark \\
        AGM \cite{AGM_Li}   & Optimazation                                       & Performance Score                                                              & $3$                                         & \checkmark    &  \checkmark & \checkmark  \\
        PMR \cite{PMR_Fan}       & Optimazation                                     & Prototype                                                       & $2$                                           & \checkmark    &  & \checkmark\\
        Relearning \cite{Relearning_wei}       & Optimazation                                     & Clustering                                                     & $3$                                          & \checkmark    & \checkmark & \checkmark \\
        ReconBoost \cite{ReconBoost_Huang}       & Optimazation                                     & Classification Loss                                                    & $3$                                          & \checkmark    & \checkmark & \checkmark \\

        \bottomrule                                
    \end{tabular}
    \vspace{-10pt}
\end{table*}

\subsection{Data}
This part focuses on the method which enhances modality performance through targeted data processing strategies. Wei et al. \cite{Sample_wei} propose a fine-grained evaluation method to facilitate multimodal collaboration. It evaluates modality-specific contributions at the sample level and employs selective resampling techniques to enhance the discriminative capabilities of weak modality modalities.

\subsection{Feed-forward}
These methods alleviate the imbalanced learning across modalities by modifying the forward process during model training and inference. These methods can be categorized into two types based on where modifications are made.

\textbf{Feature Processing.} The first type of methods adjust features during training. Adaptive Mask Co-optimization (AMCo) \cite{AMCo} masks features of dominant modalities based on their performance, while On-the-fly Prediction Modulation (OPM) \cite{OPM_PAMI} drops its feature with dynamical probability in feed-forward stage.

\textbf{Fusion Module.} The second type achieves modality balance by modifying the fusion mechanisms. Multimodal Learning with Alternating Unimodal Adaptation (MLA) \cite{MLA} uses dynamic fusion to integrate different modalities. It also employs an alternating optimization approach to optimize unimodal encoders, minimizing interference between modalities. Greedy \cite{Greedy_Wu} utilizes the MMTM \cite{MMTM_Joze} architecture for intermediate fusion to boost the modality interaction. It also facilitates the learning of weak modality that indicated by conditional learning speed, which is measured by the gradient change ratio.

\subsection{Objective}
Various methods for addressing modality imbalance in multimodal learning focus on modifying objectives. These methods can be categorized into three main directions: 

Firstly, several methods modify the multimodal loss function to mitigate the multimodal imbalance problem. For instance, Multi-Modal Cosine loss (MMCosine) \cite{MMCosine_Xu} proposes a multimodal cosine loss, which effectively increases the learning proportion of weaker modalities by weight constraints and inter-symmetric constraints.

Secondly, a group of methods leverage modality differences for learning objectives to achieve balanced learning. MBSD \cite{MBSD} constrains the model using the Kullback-Leibler (KL) \cite{KL} divergence of prediction distributions between different modalities to reduce their distance. Calibrating Multimodal Learning (CML) \cite{CML_Ma} uses confidence loss derived from different modalities, which lowers the confidence of the dominant modality. LFM \cite{LFM_yang} bridges heterogeneous data in the feature space through contrastive learning, reducing the distance between different modalities.

Thirdly, several approaches incorporate unimodal loss into the objectives to mitigate the imbalance problem. Uni-Modal Teacher (UMT) \cite{UMT_Du} introduces a unimodal distillation loss, enhancing the learning of unimodal encoders. Gradient-Blending (GBlending) \cite{Gblending_Wang} and MMPareto \cite{MMPareto_Wei} utilize unimodal losses to solve the imbalance problem. GBlending \cite{Gblending_Wang} uses overfitting-to-generalization-ratio (OGR) as an indicator to show which modality is dominant and its corresponding weight, while MMPareto \cite{MMPareto_Wei} borrows ideas from Pareto method \cite{ParetoInMultitask} to guarantee the final gradient is with direction common to all learning objectives to boost the learning of weak modality. 

\subsection{Optimization}
Recent studies have investigated optimization-based approaches to mitigate the multimodal imbalance problem. Both On-the-fly Gradient Modulation (OGM) \cite{OGM_CVPR} and Adaptive Gradient Modulation (AGM) \cite{AGM_Li} aim to balance modality learning by slowing down the gradients of dominant modalities to provide more optimization space for weak modalities. Specifically, OGM \cite{OGM_CVPR} uses performance score as an indicator to achieve this, while AGM \cite{AGM_Li} employs a Shapley value-based method for gradient adjustment. Prototypical Modality Rebalance (PMR) \cite{PMR_Fan} adjusts gradient magnitudes based on category prototypes to accelerate the learning of weak modalities. Diagnosing \& Re-learning (Relearning) \cite{Relearning_wei} uses re-initialization to reduce the dependence on dominant modalities while preventing weak modalities from learning excessive noise. ReconBoost \cite{ReconBoost_Huang} introduces an alternating-boosting optimization way to enhance the unimodal performance, which alleviates the imbalance problem.

%% file: Methods/Methods.tex
\vspace{-5pt}
\section{Toolkit}
\label{sec:tool}
To accompany BalanceBenchmark, we propose a comprehensive toolkit named \textbf{BalanceMM}, that incorporates 17 multimodal imbalance algorithms. Although these algorithms cover various methodological aspects, \textit{the toolkit provides a standardized implementation that unifies their evaluation and comparison.} Due to its modular architecture, BalanceMM allows flexible integration of various datasets, modalities, backbones and methods. This makes it extensible, allowing users to easily add new components to the overall framework. 

\subsection{Datasets and modalities} \label{sec:tool1}
BalanceMM includes 7 datasets covering multiple modalities. These datasets include both bimodal and trimodal datasets, each with varying imbalance degrees, allowing for a more comprehensive evaluation of different methods. To streamline the utilization of these datasets, we develop standardized data loaders for each dataset, ensuring consistency and reproducibility across experiments. A more detailed description of these datasets can be found in Section \ref{sec:datasets}, where we discuss their characteristics in depth.

\subsection{Backbones} \label{sec:tool2}
To provide adaptability to different modalities, BalanceMM supports alternative backbones, including ResNet18 \cite{ResNet} and Transformer \cite{Transformer}. Vision Transformer (ViT) \cite{ViT}, which serves as a variant of Transformer specifically designed for vision tasks, is also supported. Users can choose to use a backbone trained from scratch or select a pre-trained version, depending on their specific needs. Designed as a plug-and-play component, the backbone integrates seamlessly into the workflow. Moreover, the toolkit is extensible, allowing users to easily incorporate new backbones for a wide range of applications.

\subsection{Multimodal imbalance algorithms} \label{sec:tool3}
BalanceMM covers 17 multimodal imbalance algorithms spanning 4 methodological categories defined in Section \ref{sec:taxonomy}. As summarized in Table \ref{tab:mm_algorithms}, these algorithms encompass various modality combinations and application domains, such as Computer Vision (CV), Natural Language Processing (NLP), and audio. A configuration-based workflow enables the activation of any method with a single command, while maintaining the original specifications. The implementation of multimodal imbalance algorithms is illustrated in Algorithm \ref{algorithm:1}.

\subsection{Evaluation metrics} \label{sec:tool4}
BalanceMM offers unified evaluation metrics to assess multimodal imbalance methods by the criteria below.

\paragraph{Performance.} We utilize Top-1 accuracy and F1 score as our performance evaluation metrics, which are widely used in classification task.

\paragraph{Imbalance.} To quantitatively assess the degree of imbalance in multimodal learning, we introduce a metric based on the Shapley value \cite{shapley:book1952}. For a multimodal dataset with a modality set $M$, contribution $\phi^i$ for modality $i$ is computed by the Shapley value as below:
\begin{equation} \label{equ:1}
\phi^i = \frac{1}{|M|!} \sum_{\pi \in \Pi_M} \left[ v(S_\pi^i \cup \{i\}) - v(S_\pi^i) \right] ,
\end{equation}
where $\Pi_M$ denotes all permutations of $M$, $S_\pi^i$ represents the set of modalities preceding $i$ in permutation $\pi$, and $v(A)$ is the value function measuring model performance when using modality subset $A \subseteq M$. 
The value function $v(A)$ is implemented through masked evaluation, where the performance of the model is measured by the accuracy, calculated as follows:

\begin{equation} \label{equ:2}
v(A) = \frac{\sum_{k=1}^{N} \mathbf{1}(\hat{y}_k = y_k)}{N},
\end{equation}

The imbalance metric $\mathcal{I}$ is then defined as follows: for the bimodal case, 
\begin{equation} \label{equ:4}
\mathcal{I} = |\phi^1 - \phi^2|,
\end{equation} and for the trimodal case, 
\begin{equation} \label{equ:5}
\mathcal{I} = \frac{1}{3}\left(|\phi^1-\phi^2| + |\phi^1-\phi^3| + |\phi^2-\phi^3|\right).
\end{equation}
This metric satisfies three key properties:
\begin{itemize}
    \item \textbf{Null contribution}: \(\mathcal{I} = 0\) when all modalities contribute equally.
    \item \textbf{Bounded range}: \(\mathcal{I} \in [0, 1]\), following its calculation principle.
    \item \textbf{Permutation invariance}: The metric is invariant to the ordering of modalities.
\end{itemize}

This Shapley-based metric explicitly measures how much each modality contributes to the whole performance relative to other modalities. Lower $\mathcal{I}$ values indicate more balanced multimodal cooperation, while higher values suggest dominance by specific modalities.

\paragraph{Complexity.} To evaluate the computational complexity of various methods, our toolkit measures the number of \textbf{floating-point operations (FLOPs)} required during training. FLOPs represent the total number of arithmetic operations, where higher FLOPs indicate greater computational cost. This metric can help to compare efficiency of different algorithms and assess the trade-off between performance and computational overhead.

\subsection{Implementation pipeline} \label{sec:tool5}
In Algorithm \ref{algorithm:1}, we provide a reference implementation in the BalanceMM framework. The modular architecture of BalanceMM facilitates the efficient integration of various components. This not only makes the toolkit a powerful resource for evaluating multimodal imbalance algorithms, but also streamlines the experimental workflow while maintaining robust performance and adaptability.
\begin{algorithm}[t] 
\caption{The pseudo code for multimodal imbalance algorithms implementation with BalanceMM toolkit}
\begin{algorithmic} \label{algorithm:1}
\setlength{\leftskip}{0pt} 
\raggedright

\STATE \hspace*{-1em}\textbf{Input:} The selected multimodal imbalance method $\mathcal{F}$; 
\STATE \hspace*{-1em}specific hyper-parameters for the method denoted as $\alpha$;
\STATE \hspace*{-1em}the selected dataset $D$; global configuration (args).

\STATE \hspace*{-1em}\textbf{Output:} model, training logs and evaluation metrics.

\STATE \hspace*{-1em}from BalanceMM.utils.data\_utils import create\_dataloader
\STATE \hspace*{-1em}from BalanceMM.models import create\_model

\STATE \hspace*{-1em}from BalanceMM.trainers import create\_trainer

\STATE \hspace*{-1em}\textcolor{gray}{\# Load the selected dataset}
\STATE \hspace*{-1em}train\_data, val\_data, test\_data = create\_dataloader($D$)

\STATE \hspace*{-1em}\textcolor{gray}{\# Modify specific components based on method type}
\STATE \hspace*{-1em}\textbf{if} $\mathcal{F}$ in \textit{Objective} \textbf{then}
\STATE \hspace*{0em}args.trainer.loss = $L_{new}$

\STATE \hspace*{-1em}\textbf{elif} $\mathcal{F}$ in \textit{Optimization} \textbf{then}
\STATE \hspace*{0em}Set up modulation mechanism $G$ with $\alpha$ adjusting intensity of optimization
\STATE \hspace*{0em}args.trainer.modulation = $G$
\STATE \hspace*{-1em}\textbf{elif} $\mathcal{F}$ in \textit{Feed-forward} \textbf{then}
\STATE \hspace*{0em}Modify args.model.feature\_process and args.model.fusion\_module based on $\mathcal{F}$

\STATE \hspace*{-1em}\textbf{elif} $\mathcal{F}$ in \textit{Data} \textbf{then}

\STATE \hspace*{0em} args.trainer.if\_resample = True
\STATE \hspace*{-1em}model = create\_model(args.model)
\STATE \hspace*{-1em}trainer = create\_trainer(args.trainer)
\STATE \hspace*{-1em}trainer.fit(model, train\_data, val\_data)
\STATE \hspace*{-1em}performance, imbalance, complexity = 
\STATE \hspace*{-1em}trainer.evaluation(model, test\_data)
\end{algorithmic}
\end{algorithm}

%% file: Datasets/Datasets.tex
\section{Datasets and benchmark} \label{sec:datasets and benchmarks}
\subsection{Datasets} \label{sec:datasets}
BalanceBenchmark includes 7 datasets to evaluate different multimodal imbalance algorithms. These datasets include different types and numbers of modalities, as well as varying degrees of imbalance. \textbf{KineticsSounds} \cite{kinetics-sounds}, \textbf{CREMA-D} \cite{cremad}, \textbf{BalancedAV} \cite{balance}, and \textbf{VGGSound} \cite{vggsound} are audio-video datasets across various application scenarios. \textbf{UCF-101} \cite{ucf101} is a dataset with two modalities, RGB and optical flow. \textbf{FOOD-101} \cite{food101} is an image-text dataset. And \textbf{CMU-MOSEI} \cite{mosei} is a trimodal dataset (audio, video, text).

\subsection{Benchmark} \label{sec:benchmarks}
BalanceBenchmark is the first comprehensive framework designed to evaluate multimodal imbalance algorithms. It addresses three critical limitations of existing measurement approaches. Firstly, to tackle the absence of standardized metrics for imbalance analysis,  we introduce a systematic evaluation protocol in Section \ref{sec:tool4}, which measures three key dimensions in multimodal learning: performance, imbalance, and complexity. Secondly, to ensure reproducibility and fair comparison of multiple methods, we maintain consistent experimental settings through a modular toolkit with unified data loaders and backbone support. Thirdly, to prevent overfitting to specific scenarios, we incorporate 7 diverse datasets spanning different modality combinations such as audio-video, image-text, RGB-optical flow and trimodal scenarios, with varying degrees of modality imbalance.

\textbf{Implementation details.} 
To ensure a reliable comparison across methods, consistent experimental settings are maintained for each dataset. Most datasets utilize the SGD optimizer with momentum set to 0.9 and weight decay of 1e-4, while VGGSound employs an AdamW optimizer with weight decay of 1e-3. All datasets use the StepLR scheduler with a decay rate of 0.1, where the step size is 30 for most datasets and 10 for VGGSound. The batch size is fixed at 64 for most datasets, except for VGGSound which uses 32. Models on VGGSound are trained for 30 epochs, while models on other datasets are trained for 70 epochs. Learning rates are tailored to each dataset to accommodate varying training dynamics: CREMA-D, FOOD-101, KineticsSounds and VGGSound use 1e-3, BalancedAV uses 5e-3, UCF-101 and CMU-MOSEI use 1e-2. Regarding network architectures, ResNet18 is employed as the backbone for audio-video datasets (i.e., CREMA-D, KineticsSounds, BalancedAV, and VGGSound). FOOD-101 combines a pre-trained Transformer with ResNet18. UCF-101 uses ResNet18, and CMU-MOSEI applies a Transformer architecture across all three modalities. The experiments are conducted on different GPU platforms, ensuring consistency within each dataset: CREMA-D, BalancedAV, CMU-MOSEI and VGGSound are evaluated on NVIDIA GeForce RTX 3090, where VGGSound specifically uses two GPUs. KineticsSounds, FOOD-101, and UCF-101 experiments are performed on an NVIDIA A40.

%% file: Experiments/Experiments.tex
\section{Experiments and analysis} \label{sec:analysis}

\begin{table*}[!t]
\small
\caption{\textbf{Comparison of all the multimodal imbalance algorithms.} Bold and underline represent the best and second best respectively. }
\vspace{-5pt}
\setlength{\tabcolsep}{3pt}
\centering
\begin{tabular}{ccccccccccccccc}
\toprule
Method &
  \multicolumn{2}{c}{KineticsSounds} &
  \multicolumn{2}{c}{CREMA-D} &
  \multicolumn{2}{c}{UCF-101} &
  \multicolumn{2}{c}{FOOD-101} &
  \multicolumn{2}{c}{CMU-MOSEI} &
  \multicolumn{2}{c}{BalancedAV} &
  \multicolumn{2}{c}{VGGSound} \\ \cmidrule(lr){2-3} \cmidrule(lr){4-5} \cmidrule(lr){6-7} \cmidrule(lr){8-9} \cmidrule(lr){10-11} \cmidrule(lr){12-13} \cmidrule(lr){14-15}
           & ACC     & F1   & ACC     & F1   & ACC     & F1   & ACC     & F1   & ACC     & F1   & ACC     & F1   &ACC &F1\\ \midrule
Unimodal-1 & 55.06    & 54.96 & 59.38      & 59.23 & 70.55    & 69.94 & 86.19      & 86.10 & 71.09 & 41.70  & 65.34 & 62.12    & 41.27    & 40.32 \\
Unimodal-2 & 45.31    & 43.76 & 58.10      & 56.81 & 78.60    & 77.49 & 65.67      & 65.47 & 71.03  & 41.68   & 50.55 & 47.14    & 30.43 & 29.61    \\
Unimodal-3 & --    & -- & --      & -- & --    & -- & --      & -- & 80.58    & 74.57 & --    & -- & --    & -- \\
Baseline   & 65.63 & 65.28 & 65.50  & 65.07 & 81.80 & 81.21 & 91.65 & 91.60 & 78.99 & 69.40 & 73.33 & 70.73 & 48.08    & 46.98 \\ \midrule
MMCosine   & 67.49 & 67.09 & 67.19 & 67.34 & 82.97 & 82.47 & 92.16 & 92.12 & 80.38 & 73.67 & 75.05 & \underline{72.57} & 48.73    & 47.66 \\
UMT       & 68.60 & 68.43 & 67.47 & 67.75 & 84.18 & 83.56 & 93.02 & 92.96 & 80.73 & 73.60 & 74.35 & 71.68  & \textbf{51.58}    & \textbf{50.48} \\
MBSD       & 68.82 & 68.28 & 74.86 & 75.48 & 84.61 & 84.26 & \textbf{93.16} & \textbf{93.09} & 79.41 & 71.13 & 75.13 & 72.08 & 49.48    & 47.99 \\
CML        & 67.56 & 67.22 & 69.18 & 69.57 & 84.74 & 84.28 & 92.70 & 92.66 & 79.69 & 73.16 & 71.85 & 68.58 & 50.50    & 49.30 \\
GBlending  & 68.82 & 66.43 & 71.59 & 71.72 & 85.01 & 84.50 & 92.56 & 92.50 & 79.64 & 73.29 & 74.19 & 71.57 & \underline{51.41}    & \underline{50.39} \\
MMPareto   & \textbf{74.55}    & \textbf{74.21} & \textbf{79.97}    & \textbf{80.57} & 85.30    & 84.89 & 92.82    & 92.77 & \textbf{81.18 }   & \textbf{74.64} & \underline{75.26}    & 72.16 & 49.35    & 48.48 \\
LFM         & 66.37    & 66.02 & 70.02    & 69.55 & 84.95    & 84.35 & 92.58    & 92.54 & 79.90    & 71.60 & 73.82    & 70.79 & 47.45    & 46.50 \\
\textbf{Objective Avg} & 68.89 & 68.24 & 71.47 & 71.71 & 84.53 & 84.04 & 92.71 & 92.66 & 80.13 & 73.01 & 74.23 & 71.34 & 49.79 & 48.69 \\ \hline
OGM         & 67.04 & 66.95 & 67.76 & 68.02 & 82.07 & 81.30 & 91.81 & 91.77 & 80.45 & 73.61 & 73.83 & 71.49 & 48.25    & 47.16 \\
AGM        & 66.62 & 65.88 & 71.59 & 72.11 & 81.70  & 80.89 & 91.89 & 91.84 & 79.86 & 71.89 & \textbf{75.49 }& \textbf{73.09} & 49.06    & 47.70 \\
PMR        & 67.11 & 66.87 & 67.19 & 67.20 & 81.93 & 81.48 & 92.10 & 92.04 & 79.88 & 72.09 & 73.70 & 71.04 & 50.38    & 49.01 \\
Relearning  & 65.92    & 65.48 & 71.02    & 71.46 & 82.87    & 82.15 & 91.68    & 91.63 & 78.65    & 70.02 & 73.96    & 71.62 & 48.12    & 47.04 \\ 
ReconBoost  & 68.38    & 67.68 & 74.01    & 74.52 & 82.89    & 82.26 & 92.47    & 92.44 & \underline{81.01}    & \underline{74.03} & 74.66    & 72.03 & 47.27    & 46.26 \\
\textbf{Optimization Avg} & 67.01 & 66.57  & 70.31 & 70.66 & 82.29 & 81.62 & 91.99 & 91.94 & 79.97 & 72.33 & 74.33 & 71.85 & 48.62 & 47.43 \\ \midrule
MLA       & 69.05    & 68.75 & 72.30    & 72.66 & \underline{85.38}    & \underline{84.84} & \underline{93.14}    & \textbf{93.09} & 78.65    & 70.02 & 73.80    & 70.82 & 49.99    & 48.62 \\
OPM        & 66.89    & 66.44 & 68.75    & 69.00 & 85.28    & 83.79 & 93.08    & \underline{93.04} & 79.95    & 72.83 & 75.03    & 72.27 & 49.12    & 48.24 \\
Greedy    & 66.82 & 66.53 & 66.48 & 66.54 & -- & -- & -- & -- & --    & -- & 73.80 & 71.21 & 48.65    & 47.41 \\
AMCo       & \underline{70.54} & \underline{69.95} & 73.30 & 73.95 & \textbf{86.91} & \textbf{86.66} & 92.73 & 92.68 & 79.51 & 71.14 & 75.00 & 72.02 & 49.05    & 47.18 \\ 
\textbf{Forward Avg} & 68.32 & 67.91 & 70.21 & 70.54 & 85.86 & 85.10 & 92.66 & 92.94 & 79.37 & 71.33 & 74.41 & 71.58 & 49.20 & 47.86 \\ \midrule
Modality-valuation     & 68.01    & 68.03 & \underline{75.85}    & \underline{76.68} & 85.25    & 84.69 & 92.20    & 92.15 & 79.84    & 72.99 & 73.52    & 70.61 & 48.25    & 47.22 \\ \bottomrule
\end{tabular}
\vspace{-10pt}
\label{tab:experiments}
\end{table*}

                            \subsection{Experimental outcomes} 
We evaluated the effectiveness of all related methods discussed in Section \ref{sec:taxonomy}. Unimodal-1 refers to training the model using only the audio modality for KineticsSounds, CREMA-D, CMU-MOSEI, BalancedAV, and VGG. For UCF-101, it corresponds to the optical flow modality, while for FOOD-101, it refers to the text modality. Unimodal-2 refers to training the model using only the video modality for KineticsSounds, CREMA-D, CMU-MOSEI, BalancedAV, and VGG. For FOOD-101, it refers to image modality. For UCF-101, it corresponds to the RGB modality. Unimodal-3 applies only to CMU-MOSEI, where the model is trained using the text modality. Baseline refers to the commonly used approach in multimodal imbalance learning, which employs concatenation fusion with a single multimodal cross-entropy loss function. As shown in Table \ref{tab:experiments}, we conducted comprehensive experiments using the proposed BanlenceBenchmark on 7 datasets. The results indicate that almost all related methods outperform the Baseline in terms of accuracy and F1 score, demonstrating that the multimodal imbalance problem is prevalent across various scenarios. Meanwhile, addressing this problem is crucial for improving model performance. 

\begin{table}[h]
    \centering
    \small
    \setlength{\tabcolsep}{3pt}
    \caption{Average FLOPs of different categories.}
    \vspace{-5pt}
    \begin{tabular}{lcccc}
        \toprule
        Categories & Objective & Optimization & Forward & Data \\
        \midrule
        FLOPs \footnotesize{( $\times 10^{13}$ )}  & 8.38 & 16.90 & 6.94 & 11.40 \\
        \bottomrule
    \end{tabular}
    \vspace{-10pt}
    \label{tab:flops}
\end{table}

\subsection{Analysis}
\subsubsection{Comparison of different categories of methods} 
The four categories of methods exhibit different characteristics when addressing the multimodal imbalance problem. \textit{Firstly,} as shown in Table \ref{tab:experiments}, objective-based methods perform well across all datasets except BalancedAV. This is because adjusting the learning objective significantly promotes the update dynamics of the weak modalities, thus alleviating the imbalance problem. When the imbalance degree is relatively high, improving the update dynamics of the weaker modalities effectively facilitates their learning, leading to better performance of the multimodal model. However, on BalancedAV, which exhibits the lowest imbalance degree,  performance of objective-based methods is worse than that of optimization-based and forward-based methods.
\textit{Secondly,} optimization-based methods perform well on datasets with a small imbalance degree, such as BalancedAV. This is because optimization-based methods provide fine-grained adjustments over the multimodal model's learning process. When the degree of imbalance is small, these methods can more precisely balance the modalities. However, as shown in Table \ref{tab:flops}, optimization-based methods have the highest average FLOPs, which results in greater computational resource requirements. \textit{Thirdly,} forward-based methods have the smallest average FLOPs. This is because they adjust the model in terms of feature processing and fusion methods, introducing minimal additional computational overhead. For example, Greedy \cite{Greedy_Wu} employs a specific-designed fusion to address the imbalance problem. However, this  characteristic limits the applicability of forward-based methods in general frameworks.  \textit{Fourthly,} Modality-valuation \cite{Sample_wei} is the only approach that addresses the multimodal imbalance problem at the data level. It improves the quality of the training data, but also introduces relatively high computational costs. \textit{These findings suggest that no existing method achieves a satisfactory balance between performance and computational cost.}

\begin{figure}[t]
  \centering
  \begin{subfigure}[t]{0.50\linewidth}
    \centering
    \includegraphics[width=\linewidth]{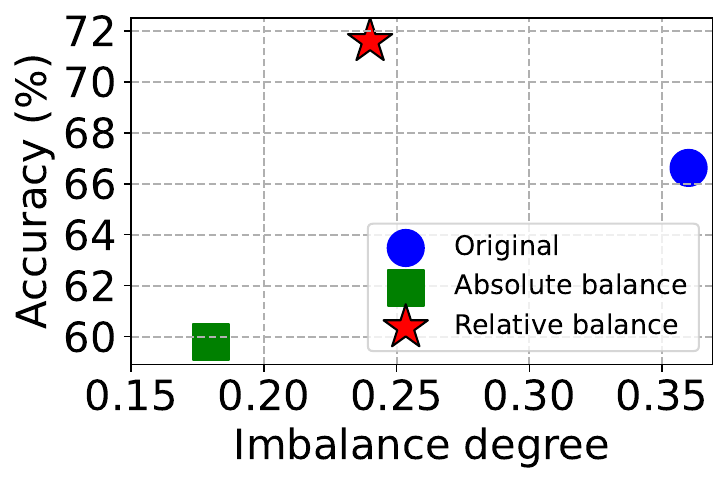}
    \vspace{-15pt}
    \caption{AGM.}
    \label{fig:AGM_Imbalance}
  \end{subfigure}%
  \hfill
  \begin{subfigure}[t]{0.50\linewidth} 
    \centering
    \includegraphics[width=\linewidth]{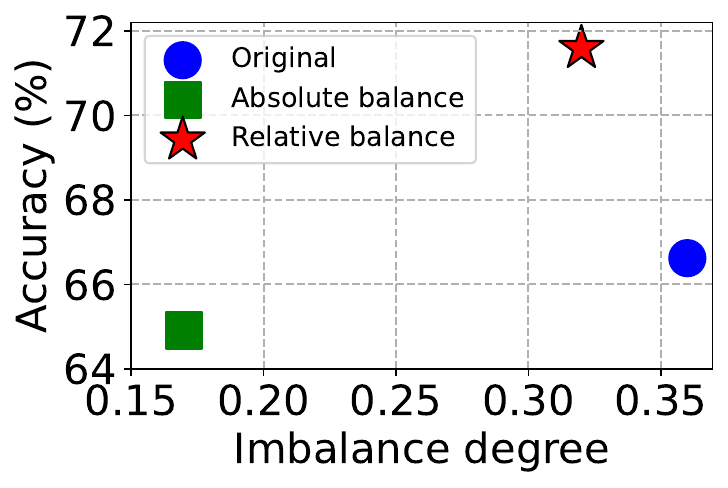}
    \vspace{-15pt}
    \caption{Gblending.}
    \label{fig:Gblending_imbalance}
  \end{subfigure}
  \vspace{-10pt}
  \caption{\textbf{(a).} Absolute and relative balance for AGM. \textbf{(b).} Absolute and relative balance for GBlending. Experiments are conducted on CREMA-D, with these two methods selected as representative cases.}
  \vspace{-15pt} 
  \label{fig:imbalance}
\end{figure}

\subsubsection{Relative balance}
We conducted comprehensive experiments to investigate the relationship between model performance and the degree of imbalance. To quantify the imbalance degree, we utilized the Shapley-based method introduced in Section \ref{sec:tool}, where higher values indicate a higher imbalance degree and lower values reflect better balance between modalities. By adjusting the hyperparameters of various methods, we obtained different combinations of imbalance degree and performance. Specifically, we identified the points with the lowest imbalance degree and the highest performance.

As illustrated in Figure \ref{fig:imbalance}, we selected visualizations from two methods to demonstrate the relationship between performance and imbalance degree. The original baseline exhibited a high imbalance degree and relatively low accuracy. Through hyperparameter tuning, we adjusted the imbalance degree between modalities and obtained varying performance results. 
When the imbalance degree is high, gradually reducing it leads to continuous performance improvement. However, once the imbalance degree reaches a relatively low level, further reduction no longer enhances performance. We refer to this point as the \textit{relative balance point}. Beyond this point, further decreasing the imbalance degree achieves the absolute balance point, where the imbalance between modalities is minimized. However, the performance at the \textit{absolute balance point} is inferior to that at the relative balance point and can even fall below the baseline. This phenomenon occurs because different modalities contain varying amounts of information. An excessive focus on balance may cause the model to undervalue high-information modalities, leading to reduced effectiveness in learning from these modalities.

\subsubsection{Future work}
Based on the analysis above, we provide several insights for future research in this field. \paragraph{Hybrid strategies.} Future research could explore hybrid strategies that integrate the strengths of different methods while mitigating their limitations. For instance, a more fine-grained adjustment of the learning objective could combine the advantages of both objective-based and optimization-based methods. 
\paragraph{Pursue relative balance.} When addressing the imbalance problem, it is important to recognize that different modalities inherently contain different amounts of information. Therefore, maintaining a relatively balanced state among modalities is preferable to blindly pursuing absolute balance. Future work could further explore efficient strategies to achieve relative balance across modalities, ensuring that models can effectively leverage the unique contributions of each modality
\paragraph{Multimodal imbalance in foundation models.} Existing methods for addressing multimodal imbalance remain limited to traditional neural networks and relatively small datasets.  However, recent studies have identified the multimodal imbalance problem in mixed-modal foundation models \cite{team2024chameleon,aghajanyan2023scaling,the2024large}. For example, studies on Chameleon \cite{team2024chameleon} shows that different modalities compete with each other with the standard LLaMA architecture \cite{touvron2023llama}. Future work could extend network architectures to foundation models.

%% file: Analysis/Analysis_Conclusions.tex
\section{Conclusion}
In conclusion, we introduce BalanceBenchmark, a unified benchmark for fair and comprehensive evaluation of multimodal imbalance algorithms. By incorporating a systematic taxonomy, diverse evaluation metrics, a comprehensive dataset collection, and the modular toolkit BalanceMM, our benchmark enables thorough assessment of existing methods and provides a convenient tool for future work. 